\documentclass{article}

\usepackage{style/arxiv}
\usepackage{subfig}
\usepackage[utf8]{inputenc} 
\usepackage[T1]{fontenc}    
\usepackage{hyperref}       
\usepackage{url}            
\usepackage{booktabs}       
\usepackage{amsfonts}       
\usepackage{nicefrac}       
\usepackage{microtype}      
\usepackage{lipsum}
\usepackage{amsmath}
\usepackage{graphicx}
\usepackage[linesnumbered,algoruled,boxed,lined]{algorithm2e}
\makeatletter 
\g@addto@macro{\@algocf@init}{\SetKwInOut{Parameter}{Parameters}} 
\makeatother
\title{G-SMOTE: A GMM-based synthetic minority oversampling technique for imbalanced learning}

\author{
  Tianlun Zhang\\
  College of Information Science and Technology,\\
  Dalian Maritime University\\
   \\
  \texttt{tlzhang@dlmu.edu.cn} \\
   \And
 Xi Yang \\
  College of Information Science and Technology,\\
  Dalian Maritime University\\
   \\
  \texttt{yokiqust@dlmu.edu.cn} \\
  \thanks{Foundation item: National Natural Science Foundation of China (61602077, 61602077); Natural Science Foundation of Liaoning Province of China under Grant (20170540097); Fundamental Research Funds for the Central Universities under Grant (3132016348)}
}

\begin{document}
\maketitle

\begin{abstract}
Imbalanced Learning is an important learning algorithm for the classification models, which have enjoyed much popularity on many applications. Typically, imbalanced learning algorithms can be partitioned into two types, i.e., data level approaches and algorithm level approaches. In this paper, the focus is to develop a robust synthetic minority oversampling technique which falls the umbrella of data level approaches. On one hand, we proposed a method to generate synthetic samples in a high dimensional feature space, instead of a linear sampling space. On the other hand, in the proposed imbalanced learning framework, Gaussian Mixture Model is employed to distinguish the outliers from minority class instances and filter out the synthetic majority class instances. Last and more importantly, an adaptive optimization method is proposed to optimize these parameters in sampling process. By doing so, an effectiveness and efficiency imbalanced learning framework is developed.
\end{abstract}


\section{Introduction}
With the great influx of attention concentrating on the classification learning, the research of imbalanced learning gradually becomes an overwhelming trend \citep{1} \citep{2} \citep{3}. In most applications \citep{4} \citep{5} \citep{6}, varying types of classifiers are employed to learn the inductive rules from a history of instances, and are then deployed to annotate label for each online instance. According to these inductive rules learned from training dataset, classification algorithms aim to provide favorable accuracies across overall categories. Ideally, most previous work are developed on two assumptions, i.e. balanced class distribution and identical misclassification cost. Consequently, these works usually fail to generalize adequate rules over the instance space when suffered from the form of imbalance.

In practice, datasets with disproportionate number of category examples commonly hinder the classification learning. To develop a classification model with favorable accuracies across overall classes in datasets, imbalanced learning is essential in this field. Typically, imbalanced learnings fall under two umbrellas, i.e. data level and algorithm level approaches \citep{7} \citep{8} \citep{9}. The data level approach often is based on sampling methods \citep{10} which modify the representative proportions of class instances in original imbalanced distribution. A well-known algorithm level method is the cost-sensitive learning \citep{11} which breaks the hypothesis of equal misclassification cost. Two types of imbalanced learnings have shown many promising benefits in most applications \citep{12}.

In this paper, the focus of our study is synthetic sampling. In regards to algorithms of synthetic sampling, the synthetic minority oversampling technique (SMOTE) is a powerful approach that has achieved a great deal of success in wide range of fields \citep{13}. The main idea of SMOTE is to create artificial minority class instances in the feature space. Though it could significantly improve classification learning, the SMOTE algorithm also has some drawbacks. First, the sampling space of SMOTE is limited in a line segment which is not reasonable for high dimensional data. On the other hand, the SMOTE cannot distinguish the outliers from minority samples, and cannot filter out the synthetic majority class instances form synthetic instances. The hybrid samples generated by SMOTE will hinder the classification learning. In our study, we will address these problems mentioned above.

In addition, a crucial issue in imbalanced learning is to assign reasonable hyper-parameters. Although there are many rules of thumb \citep{14} \citep{15}, a generic solution of this issue is necessary. We thus propose an adaptive metric in which the set of parameters associated with imbalanced learning are the objective in a process of optimization. In sum, the main topics in our study are summarized as follows.

(1) We propose an improved SMOTE that breaks the ties introduced by simple linear sampling space. The new synthetic samples generated by our proposed method have the more reasonable distribution in feature space of minority class instances.

(2) In order to address these drawbacks of SMOTE, we introduce Gaussian Mixture Model (GMM) to our proposed framework. On one hand, the GMM is employed to distinguish the outliers from minority class instances; on the other hand, synthetic majority class instances are eliminated by GMM. Comprehensive experiments prove that this proposed framework provides a more robust way to generate minority class instances.

(3) Instead of rules of thumb, an adaptive optimization method is proposed to optimize these parameters in sampling process. In this case, synthetic samples can be created in an effectiveness and efficiency way.

\section{Related work}
In this section, we will review the principal work about imbalanced learning.

Most classification learnings will fail to perform well when they suffer from complex imbalanced datasets \citep{16}. Imbalanced learning thus has high activity of advancement in various fields \citep{17} \citep{18} \citep{19} \citep{20}. Typically, there are two different categories of approaches in imbalanced learning. The first one is data level approach including random oversampling, random undersampling, synthetic minority oversampling and so on \citep{7} \citep{9} \citep{21}. Although this type of imbalanced learning is used in a wide range of applications, many studies \citep{22} \citep{23} argue that those methods can potentially depreciate classification performance because of their inherent drawbacks which cause overlapping, missing or redundant data. 

The other type of imbalanced learning is algorithm level approaches among which a popular one is cost-sensitive learning \citep{24} \citep{25} \citep{26}. Instead of adjusting original dataset to a balanced one, cost-sensitive method targets imbalanced issue by using various costs associated with different classes. Cost-sensitive is a viable learning paradigm in most cases \citep{27} \citep{28}, and draws tremendous attention. Datta et al. \citep{29} proposed near-Bayesian support vector machine (SVM) to multiclass scenario, and applied cost matrixes to imbalanced learning. Wang et al. \citep{15} investigated cost-based extreme learning machine (C-ELM) whose optimized objective is minimizing misclassifying cost. A cost based multilayer perceptron was proposed in \citep{30}, and used to two-class imbalanced learning. Bertoni et al. \citep{31} developed a semisupervised learning and used cost-sensitive neural network to graphs. Besides, some further researches have focused on within-class imbalanced problem \citep{32}.

In software engineering, imbalanced learning is a new challenge. Lamkanfi et al. \citep{33} handled imbalanced dataset, artificially. Yang et al. \citep{34} compared some imbalanced learnings in the task identifying high-impact bugs. In computer vision, Khan et al. \citep{35} proposed a convolutional neural network to tackle the imbalanced problem in image classification. However, most of the existing work are based on various empirical studies \citep{14}, an objective comparison and an adaptive process are urgently needed in practice.

\begin{figure}%
	\centering
	\subfloat[]{{\includegraphics[width=5cm]{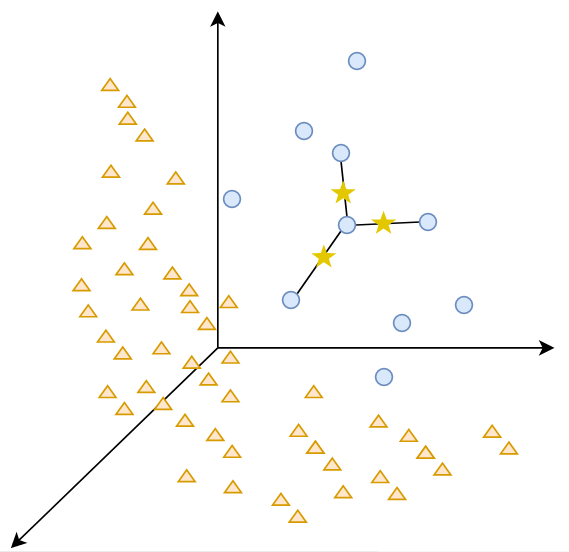} }}%
	\qquad
	\subfloat[]{{\includegraphics[width=5cm]{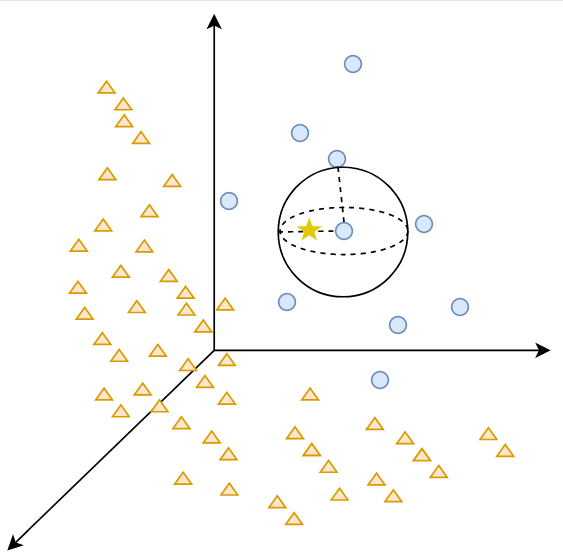} }}%
	\caption{(a) is the illustration of SMOTE; (b) shows illustration of RSMOTE.}%
	\label{fig:example}%
\end{figure}

\section{The proposed method}
\label{sec:others}
In this section, we will represent an improved SMOTE algorithm with appropriate sampling space for high dimensional data. Then the GMM-based synthetic sampling approach will be proposed. Afterward, an adaptive optimization method is proposed in our study for the hyperparameters of sampling process.

\subsection{SMOTE in high dimensional space}

To ease the presentation, some notations are established here. Suppose we have a given training dataset $S$ with $N$ cases (i.e., $|S|=N$): $S=\{(x_i,y_i)\}, i=1,2,...,N$, in which $x_i \in \mathbb{R}^{1\times n}$ is an instance in the $n$-dimensional feature space $X$, $y_i \in \{1,2,...,C\}$ is a label associated with case $x_i$. In this paper, a binary classification problem is considered, i.e. $C=2$. Two subsets are defined as $S_{min} \subset S$ and $S_{maj} \subset S$, in which $S_{min}$ denotes the the set of minority class cases, and $S_{maj}$ denotes the set of majority class cases, so that $S_{min} \cup S_{maj} = S$ and $S_{min} \cap S_{maj} = \phi$.

The synthetic minority oversampling technique (SMOTE) \citep{21} is a typical mechanism of synthetic sampling. In this algorithm, artificial cases are drawn from a feature space similarities between the instances in $S_{min}$. Concretely, in a certain neighborhood of $x_i \in S_{min}$, $K$-nearest neighbors with the smallest euclidian distance between themselves and $x_i$ are selected to create new instances for $S_{min}$. This way can be mathematically represented as follows
\begin{equation}
x_{syn}=x_i+(x_k-x_i)e
\end{equation}
where $x_k \in S_{min}$ is one of $K$-nearest neighbors $(k=1,2,3,...,K)$ of $x_i$, $e$ is a random number belonging to the range of $[0,1]$, $x_{syn}$ is the new synthetic sample of minority class. According to above equation, as shown in Figure 1(a), the star point is a new sample appearing at random location of the line joining $x_i$ and $x_k$. At first glance, this sampling approaches appear to have promising benefits because it can actually use new samples to alter the balanced degree of $S_{min}$ and $S_{maj}$. However, the SMOTE degrades the sampling space to line segments joining $x_i$ and its $K$-nearest neighbors. In practice, it is easy to extend the sampling space to the $n$-dimensional feature space $X$. In this case, the sampling space can be represented as $\Omega_{i}=\{x_{syn}| 0<|x_{syn}-x_{i}|<R\}$, where $x_i \in S_{min}$ is a sampling kernel, $R$ is the distance between $x_i$ and one of its $K$-nearest neighbors. Without loss of generality, we consider the situation of $n$=3, see Figure 1(b), synthetic samples are generated in a sphere around $x_i$. An illustration comparing different sampling methods with ours is given in Figure 2, in which one can see that the new synthetic samples generated by our proposed method have the more reasonable distribution in feature space $X$.

\begin{figure}
	\centering
	\includegraphics[width=11.5cm]{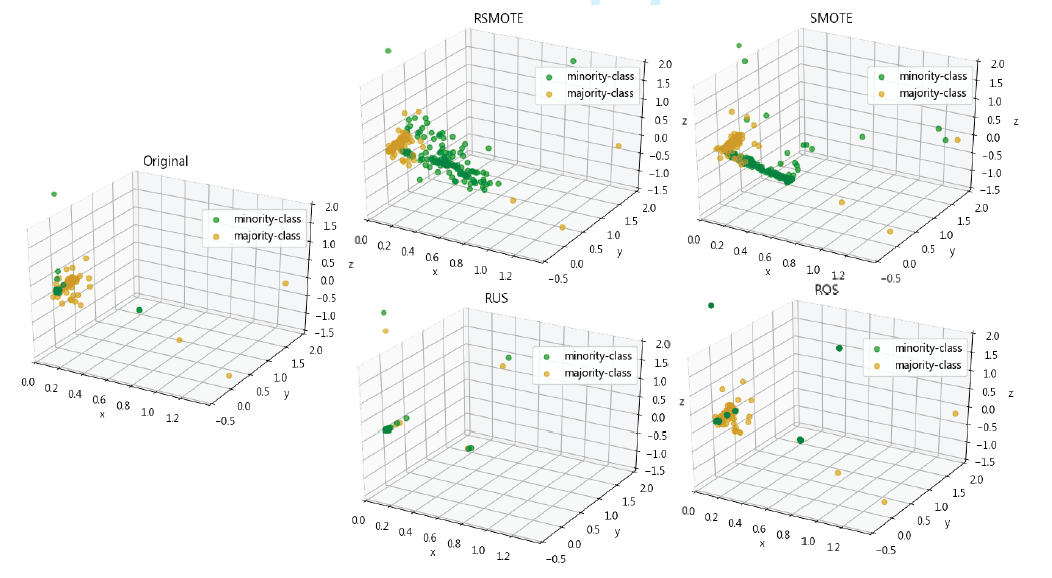}
	\caption{Comparison among RUS, ROS, SMOTE and RSMOTE on a real software engineering dataset (Core-XPConnect).}%
\end{figure}

\begin{figure}
	\centering
	\includegraphics[width=11.5cm]{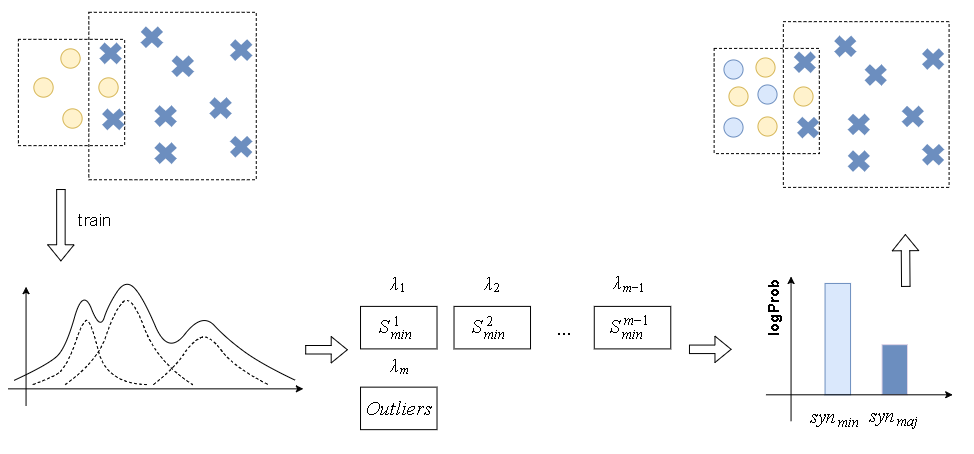}
	\caption{Entire steps of G-SMOTE.}%
\end{figure}

\subsection{GMM-based synthetic sampling approach}
This proposed method mentioned above breaks the ties introduced by simple linear sampling space. However, there still exist two obvious drawbacks in the improved SMOTE, on one hand, this method dose not have the ability to distinguish the outliers from $S_{min}$, and huge amount of synthetic outliers could lead unfavorable performance in classification learning; on the other hand, the over generalization could increase the occurrence of overlapping between classes \citep{22}, unfortunately, there not exist detection mechanism to avoid this case in the procedure of SMOTE. 

We address these problems by introducing Gaussian Mixture Model (GMM), and term this sampling method as GMM-based synthetic minority oversampling technique (GSMOTE), the entire steps of GSMOTE are illustrated in Figure fff. In the first step, the $S_{min}$ is used to train a GMM as follows

\begin{equation}
f(x|\mu,\Sigma)=\sum_{k=1}^{m}c_k\frac{1}{\sqrt{2\pi|\Sigma_{k}|}}exp[(x-\mu_{k})^T\Sigma^{-1}(x-\mu_{k})]
\end{equation}
where $c_k$, $\mu_k$ and $\Sigma_k$ respectively denote weight, mean vector and variance matrix associated with the $k$th Gaussian model component, the $m$ denotes the total number of components, the training process can be completed by expectation-maximization (EM) algorithm. After a GMM is acquired, the samples in $S_{min}$ are partitioned into $m$ cliques, see Figure 3, each $c$ presents the importance degree that the corresponding clique contributes to the overall distributive characteristics of $S_{min}$. Obviously, the clique of outliers has lower importance degree than others. Thus the instances conflicting with $S_{min}$ will be selected as sampling kernel in lower probability. 

In addition, the learned GMM is also employed to tackle the problem of over generalization. When a sampling kernel locates at the decision boundary, synthetic majority class instances will be generated in the sampling space  . It is naturally to use the learned GMM to filter out the synthetic majority class instances. As shown in Figure 3, in this step, GMM takes all of synthetic instances as input and outputs their log probability (logprob), then top-$K$ instances with the highest logprob are selected to augment the $S_{min}$. A formal description of the GSMOTE framework is shown in Algorithm 1.

\begin{algorithm}[H]
	\KwIn{
		\\ \hspace{10 mm}$m$: the number of components in GMM
		\\ \hspace{10 mm}$Num$: the number of selected sampling kernels 
		\\ \hspace{10 mm}$M$: the number of samples generated in $\Omega_i$
		\\ \hspace{10 mm}$K$: the number of instances selected from $\Omega_i$
		\\ \hspace{10 mm}$S	_{min}$: the set of minority class cases
	}
	\KwOut{
		\\ \hspace{10 mm}$S_{syn}$: the synthetic minority class samples.
	} 
	Create gaussian mixture model $gmm$ with $m$ components 
	
	Train gaussian mixture model $gmm$ on $S_{min}$
	
	Sampling $choiced\_samples$ from $S_{min}$ based on $gmm$, and $|choiced\_samples| = Num$
	
	$numattr$=Number of attributes in $choiced\_samples$
	
	$newindex$=0
	
	$synthetic$=Array[$M$ $\times$ $Num$][$numattr$]
	
	\For{$i \gets0$ \KwTo $(Num-1)$}{
		Compute $k$ nearest neighbors $nnarray$ for $choiced\_samples[i]$
		
		\For{$j \gets0$ \KwTo $(M-1)$}{
			random choice a case $nn$ from $nnarray$
			
			$r$=distance between $choiced\_samples[i]$ and $nn$
			
			Sampling a point $sample$, and distance between $sample$ and $choiced\_samples[i]$ = $d \in (0,r)$ 
			
			$synthetic[newindex]$=$sample$
			
			$newindex++$
			
		}
	}
	
	Get the top-$K$ instances $S_{syn}$ with the highest logprob from $synthetic$
	
	\caption{GSMOTE}
\end{algorithm}

\subsection{An adaptive optimization method}
One can see that there are several hyperparameters in the proposed GSMOTE framework. Instead of using the rules of thumb, we optimize these parameters by developing an adaptive optimization method, which is based on Differential evolution (DE). This iteratively searching process is given in following algorithm. For clear presentation, some notations are defined: the $Num$ denotes the number of sampling kernels selected in the first step, $M$ is the number of synthetic samples generated in $\Omega_i$, $K$ is the number of instances selected in the second step. 
\begin{table}[b]
	\caption{}
	\label{tbl:excel-table}
	\centering
	\includegraphics[width=8cm]{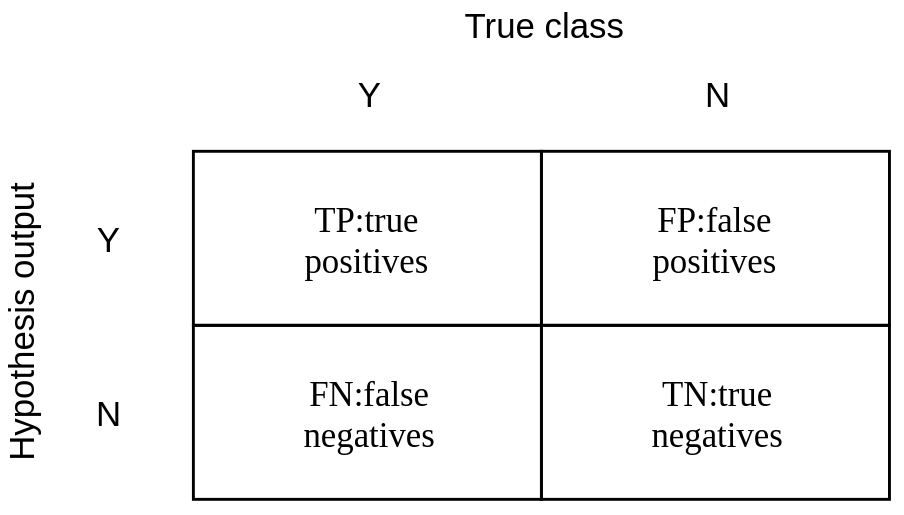}
\end{table}

\begin{algorithm}[H]
	\KwIn{
		\\ \hspace{10 mm}$G$: Maximum number of generationss
		\\ \hspace{10 mm}$N$: Population size
		\\ \hspace{10 mm}$F$: Mutation factor
		\\ \hspace{10 mm}$C_r$:Corssover probability
		\\ \hspace{10 mm}$D$: Dimension of soultion space
		\\ \hspace{10 mm}$range=(c_{min},c_{max})$: Range of values
		\\ \hspace{10 mm}$S	_{min}$: the set of minority class cases
	}
	\KwOut{
		\\ \hspace{10 mm}$c_{best}^{G}$
	}
	Generate an initial population $\{c_1^0, ..., c_N^0\}$.
	
	$c_i^0=c_{min}+rand(0,1)*(c_{max}-c_{min})$, $i=1, 2, ..., N$.
	
	Evaluate each candidate solution $c_i^0(i=1, 2, ..., N)$ in the initial population to obtain a vector representing the fitness functions $fitness(c_i^0)$ which is accuracy of ELM on dataset augmented by Algorithm 1 in this paper. In other words, $c_i^g=(m,Num,M,K)$, $g=1, 2, ..., G$.
	
	\DontPrintSemicolon
	\SetKwFunction{FMain}{fitness}
	\SetKwProg{Pn}{Function}{:}{\KwRet}
	\Pn{\FMain{$c_i^g$, $S_{min}$}}{
		$(m,Num,M,K)=c_i^g$
		
		$S_{syn}$= GRSMOTE($m,NUM,M,K,S_{min}$)
		
		$S_{aug}$ = $S$ $\cup$ $S_{syn}$
		
		train a ELM on $S_{aug}$ and calculate accuracy $acc$ on the test dataset
		
		\KwRet $acc$ \;
	}

	\For{$g \gets1$ \KwTo $G$}{
		
		\For{$i \gets1$ \KwTo $N$}{
			Random choice three vector, $c_r^1, c_r^2, c_r^3$ from $(g-1)$ generation
			
			$h_i^g = c_r^1+F\cdot(c_r^2-c_r^3)$
			
			$v_{i,j}^g$ =
				$
					\begin{cases}
					h_{i,j}^g, & $if$ \enspace rand_j \leq C_r \\
					c_{i,j}^g, & $otherwise$
					\end{cases}
					j=1, 2, ..., D.
				$
		}
		\eIf{$fitness(v_i^g) > fitness(c_i^{g-1})$}{
			$c_i^g=v_i^g$	
		}{
			$c_i^g=c_i^{g-1}$	
		}
	}
	$c_{best}^G = argmax(f(c_{i}^G)), i=1, 2, ..., N$
	\caption{Differential Evolution}
\end{algorithm}

\section{EVALUATION ON REAL-WORLD DATASETS}
\subsection{Evaluation metrics}
In this study, the predicted labels are defined as $\{Y,N\}$, a confusion matrix representing classification performance can be illustrated in Table 1.

Based on above table, the $accuracy$ can be defined as follows
\begin{equation}
Accuracy=\frac{TP+TN}{TP+TN+FP+FN}
\end{equation}
which is a simple way to describe the performance of a classifier. In addition, another popular metric is $F-measure$ defined as
\begin{equation}
F-Measure=\frac{(1+\beta^2)\cdot Recall \cdot Precision}{\beta^2\cdot Recall \cdot Precision}
\end{equation}
where $\beta$ is a coefficient of importance, $Precision$ and $Recall$ respectively are 
\begin{equation}
Precision=\frac{TP}{TP+FP}
\end{equation}
\begin{equation}
Recall=\frac{TP}{TP+FN}
\end{equation}

According to (4), we can notice that the $F$-$Measure$ is a weighted combination of $Precision$ and $Recall$, and provides insight into the functionality of a classification algorithm. Like most studies, we use an effective measure, namely $weighted$  $F$-$Measure$, to act the second metric defined as
\begin{equation}
w-F=\frac{n_1}{total}\cdot F_1 + \frac{n_2}{total}\cdot F_2
\end{equation}

where the $n_1$ is the num of the $class_1$, $n_2$ is the num of the $class_2$ and $total$ is the num of the dataset.

\subsection{Experimental Result}

In this section, we conduct several experiments to demonstrate the effectiveness of our proposed method. The datasets are blood, page, CVFTM. The imbalanced degree of these datasets is list in Table 2. Several classification algorithms are employed in our study, namely, j48, Naive bayes classifier(NB), Random tree(RT), Support vector machine(SVM). The experimental results are given in Table 3.

One can see that the relatively balanced datasets provided by our framework boost the performances of different classification learnings in most situations. 

\begin{table}[h]
	\centering
	\caption{The imbalanced degree of three datasets}
	\begin{tabular}{|l|l|l|l|}
		\hline
		DataSet & blood & page & CVFTM \\ \hline
		Imbalanced Degree     & 3.211   & 8.497 & 4.955 \\ \hline
	\end{tabular}
\end{table}

\begin{table}[h]
	\centering
	\caption{}
	\begin{tabular}{|l|l|l|l|l|}
		\hline
		\textbf{DataSet/Measure} & \textbf{j48} & \textbf{NB} & \textbf{RT} &  \textbf{SVM}\\ \hline
		boold/$Accuracy$ & 78   & 74  & 72 &  76 \\ \hline
		$$boold$_{aug}$/$Accuracy$ &   \textbf{78.6667}    &  74   &  \textbf{77.3333}   & \textbf{76.6667} \\ \hline
		boold/$F$-$Measure$ & 74.4   &  69.7  & 68.9  &  67.5 \\ \hline
		$$boold$_{aug}$/$F$-$Measure$ &   \textbf{75.4}    &  \textbf{70.9}   &  \textbf{74.4}   & \textbf{68.3} \\ \hline
		
		page/$Accuracy$ & 96.9863   & 90.7763  & 96.5297 &  94.3379 \\ \hline
		$$page$_{aug}$/$Accuracy$ &   \textbf{97.2603}    &  \textbf{92.6941}   &  \textbf{97.3516}   & 92.0548 \\ \hline
		page/$F$-$Measure$ & 97.0   &  90.4  & 96.5  &  93.3 \\ \hline
		$$page$_{aug}$/$F$-$Measure$ &  \textbf{97.2}   &  \textbf{91.6}  &  \textbf{97.3}    & 89.6 \\ \hline
		
		CVFTM/$Accuracy$ & 99.9537   & 90.1852  & 97.2222 &  99.9537 \\ \hline
		$$CVFTM$_{aug}$/$Accuracy$ &   99.9537    &  \textbf{99.9537}   &  \textbf{99.9074}   & 99.9537 \\ \hline
		CVFTM/$F$-$Measure$ & 100.0   &  88.2  & 97.2  &  100.0 \\ \hline
		$$CVFTM$_{aug}$/$F$-$Measure$ &  100.0   &  \textbf{100.0}  &  \textbf{99.9}    & 100.0 \\ \hline
	\end{tabular}
\end{table}

Besides, some experiments are conducted on bug triaging system to recognize the severity of bugs. In this situation, classification algorithms aim to distinguish $non$-$severe$ from $severe$ bugs which a software developer must fix as soon as possible. In this study, bug reports are from three major open-source projects, i.e. Eclipse \citep{36}, Mozilla \citep{37}, and Gnome \citep{38}. Bugzilla is the common bug tracking system used by these projects. To provide input for classification algorithms, the textual strings of report are transformed to digital vectors by the preprocessing steps which will be further described in what follows.

\paragraph{Tokenization.} First of all, a large textual string is divided into a group of tokens, and each of these corresponds to a single term. Meanwhile, all meaningless words (e.g. commas and punctuations) are filtered out during this process. In this step, all capitals also are replaced by corresponding lower-case letters. 

\paragraph{Stop-words removal.} In a bug report, the stop-words, such as `in', `that' and `the', do not include much specific contextual information, however, the frequency of these symbols are higher than others. To decrease the dimensionality and redundancy of transformed digital vectors, it is essential to remove all stop-words from each token based on the known list of stop-words.

\paragraph{Stemming.} In human languages, different terms commonly carry the same contextual information, and share the same morphological base, e.g., `computerize' and `computerized' have the same basic form: `computer'. To reduce the variety of descriptions, this step maps all terms with the same specific information to their common form.

\paragraph{Term frequency.} Based above all, a single keyword vector is extracted from a single bug report by using a keyword dictionary, and a weighting method is needed in this step, i.e. $TF \times IDF$ approach. Let $D=\{d_1, ..., d_j, ..., d_m\}$ denote a set of documents, in which $d_j=\{t_1, ..., t_k\}$ is a group of terms, then the term frequency ($TF_{i,j}$) of $i$th term $t_i$ in $j$th document $d_j$ can be defined as follows:
\begin{equation}
TF_{i,j}=\frac{rf_{i,j}}{\sum_{k} rf_{k,j}}
\end{equation}
where $rf_{i,j}$ is the amount of occurrences of the $t_i$ $d_j$, and $k$ is the total number of terms in $d_j$. The other factor is inverse document frequency ($IDF_i$) representing the importance of $i$th term, which is defined as follows:
\begin{equation}
IDF_i=\log\frac{m}{df_i}
\end{equation}
where $df_i$ is the document frequency, i.e. the number of documents containing $t_i$, $m$ is the total number of documents. The importance of a particular term will subsequently decrease when this term appears in many documents.

As discussed above, the preprocessing steps take textual strings as input and outputs keyword vectors for the next step, i.e. classification learning, in which a history of reports with known severity are the training dataset. As shown in Table 5, the experimental results show that the proposed method dose improve the performances of different classifiers in bug triaging systems.

\begin{table}[h]
	\centering
	\caption{Bug report datasets}
	\begin{tabular}{|l|l|l|l|l|l|}
		\hline
		\textbf{Num} & \textbf{Product} & \textbf{Name} & \textbf{Severe} &  \textbf{Non-severe} & \textbf{Degree} \\ \hline
		1 & GNOME  & Evolution\_Contacts & 1071 & 384 & 2.789 \\ \hline
		2 & Eclipse  & CDT\_cdt-core & 273 & 66 & 4.136 \\ \hline		
		3 & Eclipse  & JDT\_Core & 789 & 306 & 2.578 \\ \hline		
		4 & Moizlla  & Core\_Printing & 702 & 99 & 7.091 \\ \hline		
	\end{tabular}
\end{table}

\begin{table}[h]
	\centering
	\caption{}
	\begin{tabular}{|l|l|l|l|l|}
		\hline
		\textbf{DataSet/Measure} & \textbf{j48} & \textbf{NB} & \textbf{RT} &  \textbf{SVM}\\ \hline
		1/$Accuracy$ & 74.089  & 76.289  & 75.739 &  76.289 \\ \hline
		$$1$_{aug}$/$Accuracy$ &   \textbf{77.113}    &  \textbf{76.770}   &  \textbf{76.632}   & \textbf{77.388} \\ \hline
		1/$F$-$Measure$ & 71.245   &  76.799  & 75.402 & 75.646 \\ \hline
		$$1$_{aug}$/$F$-$Measure$ &   \textbf{75.103}    &  \textbf{77.058}   &  \textbf{75.694}   & \textbf{77.341} \\ \hline
		
		2/$Accuracy$ & 79.941  & 66.667  & 73.451 &  76.991\\ \hline
		$$2$_{aug}$/$Accuracy$ &   \textbf{81.416}    &  \textbf{81.711}   &  \textbf{82.301}   & \textbf{77.581} \\ \hline
		2/$F$-$Measure$ & 71.554   &  69.263  & 72.208 & 74.455 \\ \hline
		$$2$_{aug}$/$F$-$Measure$ &   \textbf{76.004}    &  \textbf{76.562}   &  \textbf{79.959}   & \textbf{76.864} \\ \hline
		
		3/$Accuracy$ & 77.443  & 71.963  & 73.516 &  74.247\\ \hline
		$$3$_{aug}$/$Accuracy$ &   \textbf{77.991}    &  \textbf{77.443}   &  \textbf{76.073}   & \textbf{75.151} \\ \hline
		3/$F$-$Measure$ & 74.113   &  72.827  & 73.049 & 72.166 \\ \hline
		$$3$_{aug}$/$F$-$Measure$ &   \textbf{76.398}    &  \textbf{76.771}   &  \textbf{73.656}   & \textbf{73.403} \\ \hline
		
		4/$Accuracy$ & 86.642  & 78.527  & 84.894 &  85.893\\ \hline
		$$4$_{aug}$/$Accuracy$ &   86.642    &  \textbf{88.764}   &  \textbf{87.016}   & \textbf{84.401} \\ \hline
		4/$F$-$Measure$ & 81.586  &  80.890 & 83.456 & 83.046 \\ \hline
		$$4$_{aug}$/$F$-$Measure$ &   81.586    &  \textbf{84.942}   &  \textbf{85.457}   & 81.359 \\ \hline
	\end{tabular}
\end{table}

\bibliographystyle{abbrvnat}
\bibliography{references}

\begin{thebibliography}{38}
\providecommand{\natexlab}[1]{#1}
\providecommand{\url}[1]{\texttt{#1}}
\expandafter\ifx\csname urlstyle\endcsname\relax
  \providecommand{\doi}[1]{doi: #1}\else
  \providecommand{\doi}{doi: \begingroup \urlstyle{rm}\Url}\fi

\bibitem[Bertoni et~al.(2011)Bertoni, Frasca, and Valentini]{31}
A.~Bertoni, M.~Frasca, and G.~Valentini.
\newblock Cosnet: a cost sensitive neural network for semi-supervised learning
  in graphs.
\newblock In \emph{Joint European Conference on Machine Learning and Knowledge
  Discovery in Databases}, pages 219--234. Springer, 2011.

\bibitem[Bugzilla(2018{\natexlab{a}})]{36}
Bugzilla.
\newblock Eclipse.
\newblock \url{http://bugs.eclipse.org/bugs}, 2018{\natexlab{a}}.
\newblock Accessed: 2018-02-02.

\bibitem[Bugzilla(2018{\natexlab{b}})]{37}
Bugzilla.
\newblock Mozilla.
\newblock \url{http://bugzilla.mozilla.org}, 2018{\natexlab{b}}.
\newblock Accessed: 2018-02-02.

\bibitem[Bugzilla(2018{\natexlab{c}})]{38}
Bugzilla.
\newblock Gnome.
\newblock \url{http://bugzilla.gnome.org}, 2018{\natexlab{c}}.
\newblock Accessed: 2018-02-02.

\bibitem[Castro and Braga(2013)]{30}
C.~L. Castro and A.~P. Braga.
\newblock Novel cost-sensitive approach to improve the multilayer perceptron
  performance on imbalanced data.
\newblock \emph{IEEE transactions on neural networks and learning systems},
  24\penalty0 (6):\penalty0 888--899, 2013.

\bibitem[Chan and Stolfo(1998)]{17}
P.~K. Chan and S.~J. Stolfo.
\newblock Toward scalable learning with non-uniform class and cost
  distributions: A case study in credit card fraud detection.
\newblock In \emph{KDD}, volume~98, pages 164--168, 1998.

\bibitem[Chawla et~al.(2004{\natexlab{a}})Chawla, Japkowicz, and Kotcz]{16}
N.~Chawla, N.~Japkowicz, and A.~Kotcz.
\newblock Editorial: special issue on learning from imbalanced data sets.
  sigkdd explor newsl 6: 1--6, 2004{\natexlab{a}}.

\bibitem[Chawla et~al.(2004{\natexlab{b}})Chawla, Japkowicz, and Kotcz]{3}
N.~Chawla, N.~Japkowicz, and A.~Kotcz.
\newblock Editorial: special issue on learning from imbalanced data sets.
  sigkdd explor newsl 6: 1--6, 2004{\natexlab{b}}.

\bibitem[Chawla et~al.(2002)Chawla, Bowyer, Hall, and Kegelmeyer]{21}
N.~V. Chawla, K.~W. Bowyer, L.~O. Hall, and W.~P. Kegelmeyer.
\newblock Smote: synthetic minority over-sampling technique.
\newblock \emph{Journal of artificial intelligence research}, 16:\penalty0
  321--357, 2002.

\bibitem[Chen and Liu(2018)]{6}
C.~P. Chen and Z.~Liu.
\newblock Broad learning system: an effective and efficient incremental
  learning system without the need for deep architecture.
\newblock \emph{IEEE transactions on neural networks and learning systems},
  29\penalty0 (1):\penalty0 10--24, 2018.

\bibitem[Datta and Das(2015)]{29}
S.~Datta and S.~Das.
\newblock Near-bayesian support vector machines for imbalanced data
  classification with equal or unequal misclassification costs.
\newblock \emph{Neural Networks}, 70:\penalty0 39--52, 2015.

\bibitem[Domingos(1999)]{25}
P.~Domingos.
\newblock Metacost: A general method for making classifiers cost-sensitive.
\newblock In \emph{Proceedings of the fifth ACM SIGKDD international conference
  on Knowledge discovery and data mining}, pages 155--164. ACM, 1999.

\bibitem[Fan et~al.(1999)Fan, Stolfo, Zhang, and Chan]{28}
W.~Fan, S.~J. Stolfo, J.~Zhang, and P.~K. Chan.
\newblock Adacost: misclassification cost-sensitive boosting.
\newblock In \emph{Icml}, pages 97--105, 1999.

\bibitem[He and Garcia(2008)]{13}
H.~He and E.~A. Garcia.
\newblock Learning from imbalanced data.
\newblock \emph{IEEE Transactions on Knowledge \& Data Engineering}, \penalty0
  (9):\penalty0 1263--1284, 2008.

\bibitem[Holte et~al.(1989)Holte, Acker, Porter, et~al.]{23}
R.~C. Holte, L.~Acker, B.~W. Porter, et~al.
\newblock Concept learning and the problem of small disjuncts.
\newblock In \emph{IJCAI}, volume~89, pages 813--818. Citeseer, 1989.

\bibitem[Japkowicz(2003)]{2}
N.~Japkowicz.
\newblock Class imbalances: are we focusing on the right issue.
\newblock In \emph{Workshop on Learning from Imbalanced Data Sets II}, volume
  1723, page~63, 2003.

\bibitem[Japkowicz and Stephen(2002)]{7}
N.~Japkowicz and S.~Stephen.
\newblock The class imbalance problem: A systematic study.
\newblock \emph{Intelligent data analysis}, 6\penalty0 (5):\penalty0 429--449,
  2002.

\bibitem[Japkowicz et~al.(2000)]{1}
N.~Japkowicz et~al.
\newblock Learning from imbalanced data sets: a comparison of various
  strategies.
\newblock In \emph{AAAI workshop on learning from imbalanced data sets},
  volume~68, pages 10--15. Menlo Park, CA, 2000.

\bibitem[Jo and Japkowicz(2004)]{32}
T.~Jo and N.~Japkowicz.
\newblock Class imbalances versus small disjuncts.
\newblock \emph{ACM Sigkdd Explorations Newsletter}, 6\penalty0 (1):\penalty0
  40--49, 2004.

\bibitem[Khan et~al.(2017)Khan, Hayat, Bennamoun, Sohel, and Togneri]{35}
S.~H. Khan, M.~Hayat, M.~Bennamoun, F.~A. Sohel, and R.~Togneri.
\newblock Cost-sensitive learning of deep feature representations from
  imbalanced data.
\newblock \emph{IEEE transactions on neural networks and learning systems},
  2017.

\bibitem[Kubat et~al.(1998)Kubat, Holte, and Matwin]{20}
M.~Kubat, R.~C. Holte, and S.~Matwin.
\newblock Machine learning for the detection of oil spills in satellite radar
  images.
\newblock \emph{Machine learning}, 30\penalty0 (2-3):\penalty0 195--215, 1998.

\bibitem[Lamkanfi et~al.(2010)Lamkanfi, Demeyer, Giger, and Goethals]{33}
A.~Lamkanfi, S.~Demeyer, E.~Giger, and B.~Goethals.
\newblock Predicting the severity of a reported bug.
\newblock In \emph{Mining Software Repositories (MSR), 2010 7th IEEE Working
  Conference on}, pages 1--10. IEEE, 2010.

\bibitem[Laurikkala(2001)]{10}
J.~Laurikkala.
\newblock Improving identification of difficult small classes by balancing
  class distribution.
\newblock In \emph{Conference on Artificial Intelligence in Medicine in
  Europe}, pages 63--66. Springer, 2001.

\bibitem[Liu and Zhou(2006)]{26}
X.-Y. Liu and Z.-H. Zhou.
\newblock The influence of class imbalance on cost-sensitive learning: An
  empirical study.
\newblock In \emph{null}, pages 970--974. IEEE, 2006.

\bibitem[Phua et~al.(2004)Phua, Alahakoon, and Lee]{18}
C.~Phua, D.~Alahakoon, and V.~Lee.
\newblock Minority report in fraud detection: classification of skewed data.
\newblock \emph{Acm sigkdd explorations newsletter}, 6\penalty0 (1):\penalty0
  50--59, 2004.

\bibitem[Prati et~al.(2004)Prati, Batista, and Monard]{22}
R.~C. Prati, G.~E. Batista, and M.~C. Monard.
\newblock Class imbalances versus class overlapping: an analysis of a learning
  system behavior.
\newblock In \emph{Mexican international conference on artificial
  intelligence}, pages 312--321. Springer, 2004.

\bibitem[Sun et~al.(2007)Sun, Kamel, Wong, and Wang]{11}
Y.~Sun, M.~S. Kamel, A.~K. Wong, and Y.~Wang.
\newblock Cost-sensitive boosting for classification of imbalanced data.
\newblock \emph{Pattern Recognition}, 40\penalty0 (12):\penalty0 3358--3378,
  2007.

\bibitem[Tang et~al.(2016)Tang, Deng, and Huang]{4}
J.~Tang, C.~Deng, and G.-B. Huang.
\newblock Extreme learning machine for multilayer perceptron.
\newblock \emph{IEEE transactions on neural networks and learning systems},
  27\penalty0 (4):\penalty0 809--821, 2016.

\bibitem[Ting(2002)]{27}
K.~M. Ting.
\newblock An instance-weighting method to induce cost-sensitive trees.
\newblock \emph{IEEE Transactions on Knowledge and Data Engineering},
  14\penalty0 (3):\penalty0 659--665, 2002.

\bibitem[Wang and Japkowicz(2004)]{9}
B.~Wang and N.~Japkowicz.
\newblock Imbalanced data set learning with synthetic samples.
\newblock In \emph{Proc. IRIS Machine Learning Workshop}, volume~19, 2004.

\bibitem[Weiss(2009)]{12}
G.~M. Weiss.
\newblock Mining with rare cases.
\newblock In \emph{Data mining and knowledge discovery handbook}, pages
  747--757. Springer, 2009.

\bibitem[Weiss and Provost(2003)]{8}
G.~M. Weiss and F.~Provost.
\newblock Learning when training data are costly: The effect of class
  distribution on tree induction.
\newblock \emph{Journal of Artificial Intelligence Research}, 19:\penalty0
  315--354, 2003.

\bibitem[Woods et~al.(1993)Woods, Doss, Bowyer, Solka, Priebe, and
  KEGELMEYER~JR]{19}
K.~S. Woods, C.~C. Doss, K.~W. Bowyer, J.~L. Solka, C.~E. Priebe, and W.~P.
  KEGELMEYER~JR.
\newblock Comparative evaluation of pattern recognition techniques for
  detection of microcalcifications in mammography.
\newblock \emph{International Journal of Pattern Recognition and Artificial
  Intelligence}, 7\penalty0 (06):\penalty0 1417--1436, 1993.

\bibitem[Yang et~al.(2017)Yang, Lo, Xia, Huang, and Sun]{34}
X.-L. Yang, D.~Lo, X.~Xia, Q.~Huang, and J.-L. Sun.
\newblock High-impact bug report identification with imbalanced learning
  strategies.
\newblock \emph{Journal of Computer Science and Technology}, 32\penalty0
  (1):\penalty0 181--198, 2017.

\bibitem[Zadrozny et~al.(2003)Zadrozny, Langford, and Abe]{24}
B.~Zadrozny, J.~Langford, and N.~Abe.
\newblock Cost-sensitive learning by cost-proportionate example weighting.
\newblock In \emph{Data Mining, 2003. ICDM 2003. Third IEEE International
  Conference on}, pages 435--442. IEEE, 2003.

\bibitem[Zhou and Feng(2017)]{5}
Z.-H. Zhou and J.~Feng.
\newblock Deep forest: Towards an alternative to deep neural networks.
\newblock \emph{arXiv preprint arXiv:1702.08835}, 2017.

\bibitem[Zhu and Wang(2017)]{15}
H.~Zhu and X.~Wang.
\newblock A cost-sensitive semi-supervised learning model based on uncertainty.
\newblock \emph{Neurocomputing}, 251:\penalty0 106--114, 2017.

\bibitem[Zong et~al.(2013)Zong, Huang, and Chen]{14}
W.~Zong, G.-B. Huang, and Y.~Chen.
\newblock Weighted extreme learning machine for imbalance learning.
\newblock \emph{Neurocomputing}, 101:\penalty0 229--242, 2013.

\end{thebibliography}

\end{document}